\documentclass{article}

\usepackage{microtype}
\usepackage{graphicx}
\usepackage{subfigure}
\usepackage{booktabs} 

\usepackage{hyperref}

\usepackage[accepted]{icml2025}

\usepackage{amsmath}
\usepackage{amssymb}
\usepackage{mathtools}
\usepackage{amsthm}

\usepackage[capitalize,noabbrev]{cleveref}

\theoremstyle{plain}

\theoremstyle{definition}

\theoremstyle{remark}

\newcommand{\Sref}[1]{\S\ref{#1}}

\definecolor{purp}{HTML}{791f87}

\usepackage[textsize=tiny]{todonotes}

\icmltitlerunning{Multi-LLM Collaboration}

\begin{document}

\twocolumn[
\icmltitle{When One LLM Drools, Multi-LLM Collaboration Rules}



\icmlsetsymbol{equal}{*}

\begin{icmlauthorlist}
\icmlauthor{Shangbin Feng}{a1}
\icmlauthor{Wenxuan Ding}{a2}
\icmlauthor{Alisa Liu}{a1}
\icmlauthor{Zifeng Wang}{a3}
\icmlauthor{Weijia Shi}{a1}
\icmlauthor{Yike Wang}{a1}
\icmlauthor{Shannon Zejiang Shen}{a4}
\icmlauthor{Xiaochuang Han}{a1}
\icmlauthor{Hunter Lang}{a4}
\icmlauthor{Chen-Yu Lee}{a3}
\icmlauthor{Tomas Pfister}{a3}
\icmlauthor{Yejin Choi}{a5}
\icmlauthor{Yulia Tsvetkov}{a1}
\end{icmlauthorlist}

\icmlaffiliation{a1}{University of Washington}
\icmlaffiliation{a2}{The University of Texas at Austin}
\icmlaffiliation{a3}{Google}
\icmlaffiliation{a4}{Massachusetts Institute of Technology}
\icmlaffiliation{a5}{Stanford University}


\icmlcorrespondingauthor{Shangbin Feng}{shangbin@cs.washington.edu}


\vskip 0.3in
]



\printAffiliationsAndNotice{}  

\begin{abstract}
    This position paper argues that in many realistic (i.e., complex, contextualized, subjective) scenarios, one LLM is not enough to produce a reliable output. We challenge the status quo of relying solely on a single general-purpose LLM and argue for \emph{multi-LLM collaboration} to better represent the extensive diversity of data, skills, and people. We first posit that a single LLM underrepresents real-world data distributions, heterogeneous skills, and pluralistic populations, and that such representation gaps cannot be trivially patched by further training a single LLM. We then organize existing multi-LLM collaboration methods into a hierarchy, based on the level of access and information exchange, ranging from API-level, text-level, logit-level, to weight-level collaboration. Based on these methods, we highlight how multi-LLM collaboration addresses challenges that a single LLM struggles with, such as reliability, democratization, and pluralism. Finally, we identify the limitations of existing multi-LLM methods and motivate future work. We envision multi-LLM collaboration as an essential path toward compositional intelligence and collaborative AI development.
\end{abstract}

\section{Introduction}
\vspace{5pt}

The successes of scaling models \citep{kaplan2020scaling} and data \citep{hoffmann2022training} have fueled the overly optimistic hope that simply building an ever-larger language model is a path to achieving human-like intelligent AI models.
From research artifacts to user-facing products, the commercialization of LLM and AI technologies further reinforces this status quo: major players train a single general-purpose in-house LLM and compete by attempting to outrank other models \citep{bigtech}. This quest for the ``best'' single LLM---measured by leaderboard scores, user adoption, and profitability---has brought about the bloom of LLM research and development where new models emerge daily and the state-of-the-art is constantly reshaped \citep{liangholistic, chiangchatbot, guo2025deepseek}.

\begin{figure}
    \centering
    \includegraphics[width=0.95\linewidth]{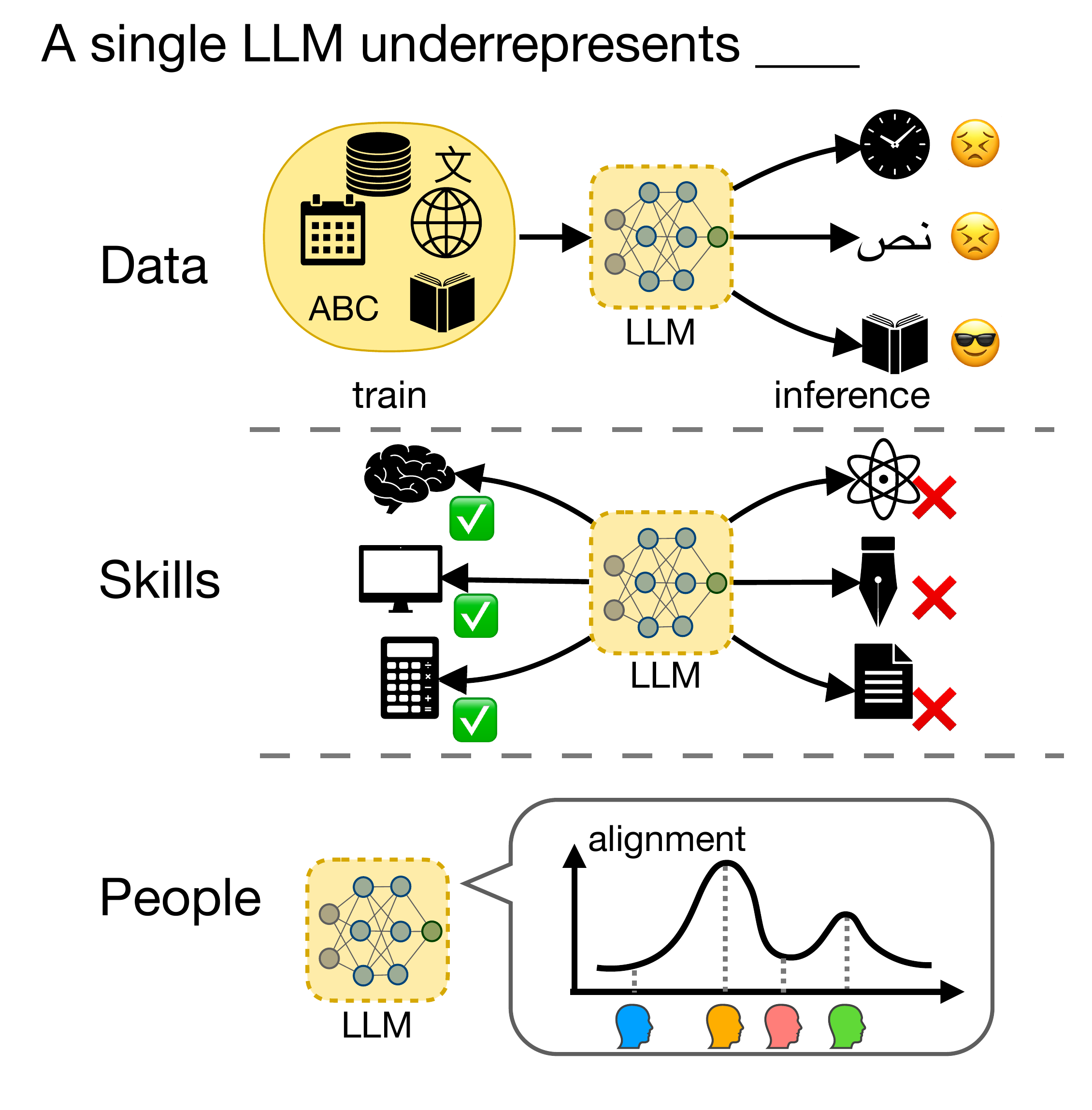}
    \vspace*{-20pt}
    \caption{Despite the quest for general-purpose models, a single LLM suffers from underrepresentation of data (language varieties, domains, styles), skills (reasoning abilities, linguistic and communication skills, creative capacities, and technical competencies), and people (opinions, values, cultural norms).
    }
    \label{fig:teaser}
\end{figure}

In this position paper, we challenge the status quo by arguing that \textbf{one LLM is not enough} and advocate for \textbf{multi-LLM collaboration}, where multiple language models collaborate for compositional generative modeling. We first argue \emph{why} one LLM is not enough (\Sref{sec:why}): despite being general-purpose, a single monolithic model struggles to reflect the intricate diversity of the real world and \emph{underrepresents} the long tail of data types, model skills, and people. We then propose a taxonomy of multi-LLM collaboration protocols (\Sref{sec:how}) in which LLMs collaborate, interact, and exchange information at the API-level, text-level, logit-level, and weight-level, offering diverse modes of collaboration compatible with all stages of the LLM lifecycle and usage types. We then argue that multi-LLM systems empowered by these protocols bring out benefits that a single LLM struggles to reflect (\Sref{sec:what}): pluralism, democratization, efficiency, adaptability, and more. Together, these arguments demonstrate that multi-LLM collaboration is an important yet overlooked family of methods, and a promising approach to advance language technologies.

We also identify some limitations of existing multi-LLM collaboration protocols and applications (\Sref{sec:discussion}), which motivate us to lay out an actionable roadmap for future work beyond monolithic models and towards advancing modular multi-LLM systems. We hope that this position will be a call-to-action for the research community to propose, evaluate, and promote collaboration strategies and communication protocols for multi-LLM collaboration.

\section{One LLM Is Not Enough}
\label{sec:why}
\vspace{5pt}

From the early successes of scaling up model size \citep{kaplan2020scaling} and training data \citep{hoffmann2022training}, language technologies have transitioned from task-specific systems to ``general-purpose''  language models \citep{Brown2020LanguageMA}: by pretraining on gigantic corpora and aligning with extensive instruction tuning and preference optimization, one LLM can be prompted to solve a diverse range of tasks and problems, leading some to believe that the future of language technologies is in figuring out the recipe for scaling and developing a single omnipotent LLM. Despite its promise, we argue that a single LLM, as designed today, is not enough to achieve a truly reliable system: even with the best effort to curate data, design architectures, and improve model inference, a single LLM suffers from \emph{underrepresentation} on three fronts: data, skills, and people.

\paragraph{Underrepresentation of data.} Despite extensive data curation efforts, a single LLM is ultimately trained on a static snapshot of what is readily available, and there are always elements in the real-world language distributions that are missing or down-weighted in this static snapshot \citep{lazaridou2021pitfalls}. For example, constant changes in the state of the world after the time of data collection quickly make the parametric information of LLMs outdated \citep{dhingra2022time, kasai2024realtime}. Private and copyrighted texts would require careful consideration in LLM training, but are otherwise essential for personalization and context \citep{weievaluating, chen2024copybench}.  Languages, dialects, language varieties in the long tail of data distributions are easily outnumbered and overshadowed by the majority languages/dialects \citep{song2023globalbench, faisal-etal-2024-dialectbench}. Evolving trends, unspoken cultural and social norms essential for socially-aware LLM applications, commonsense and implicit knowledge are hard to pin down with static data snapshots \citep{rao2024normad, shi2024culturebank}. The list goes on, and much of the real-world variation expressed through language will inevitably be lost when we solely rely on a single LLM with a static hodgepodge of training corpora.

\paragraph{Underrepresentation of skills.} Earlier language technologies were defined by task-specific progress with specialized methods, models, and subcommunities of experts for tasks like machine translation, summarization, question answering, and natural language inference \citep{sun2022paradigm}. LLMs broke from this trend by being seemingly ``general-purpose'' and it appears plausible that all we will need in the near future is a single omnipotent LLM that works best in any task and context.

However, no single LLM is Pareto-optimal \emph{empirically} and it is prohibitively expensive (if not impossible) to optimize for a single model that outperforms all other models on \emph{all} skills. For example, Gemini \citep{team2023gemini} currently ranks best on Chatbot Arena \citep{chiangchatbot} focusing on instruction following, GPT-4o \citep{achiam2023gpt} is best on the HELM leaderboard \citep{liangholistic} with an emphasis on QA and math reasoning, while a fine-tuned version of InternLM \citep{team2023internlm} is best on textual and algorithmic tasks in Big-Bench Hard \citep{suzgun2023challenging} on Open LLM Leaderboard \citep{open-llm-leaderboard-v2}.\footnote{Leaderboards accessed on Nov 24, 2024.} These models would all rankly poorly on GlobalBench \citep{song2023globalbench} and DialectBench \citep{faisal-etal-2024-dialectbench} compared to multilingual LLMs, where tasks include languages and language varieties not captured in the most popular leaderboards. This demonstrates that even the most advanced LLMs have major limitations in skills and task coverage, and that additional specialization of models is critical.


\paragraph{Underrepresentation of people.} All LLMs are ultimately used by people with diverse needs, pluralistic values, and varying socio-cultural backgrounds. Despite the ever-increasing model size and benchmark scores, we witness a constant lack of representation of actual LLM users.

On one hand, a single LLM struggles to reflect pluralistic human values, cultures, and social contexts \citep{sorensenposition, feng2024modular, leibo2024theory}, in any language. LLM users are not homogeneous, bringing a wealth of perspectives and diversity that reflects and shapes our world: despite the potential diversity in data sources, even state-of-the-art LLMs cannot equitably serve the entire spectrum of users by reflecting such heterogeneity. For example, LLMs often feature a West-centric cultural persona \citep{naous2023having} and struggle to adapt to cultural variation \citep{rao2024normad}; a single LLM would most likely reinforce the majority class in training data and exhibit biases in opinions and perspectives \citep{santurkar2023whose, feng2023pretraining}; user agency often remains overlooked since monolithic LLMs lack steerability and controllability in value-laden instructions and contexts \citep{sorensen2024value}. Since LLMs are already trained on diverse texts from the web, representing populations would require solutions beyond scaling data for a general-purpose LLM.

Moreover, by solely relying on one single model we are also solely relying on only one team of model developers. With the increasing cost and opaqueness of independently developing an LLM, these teams are becoming highly homogeneous: big tech companies, researchers with advanced degrees, overrepresentation of certain demographic groups are common sketches of the teams behind state-of-the-art LLMs \citep{workforce}. However, this leaves the vast majority of actual and underprivileged LLM users without a say in the decision making of model training and development, while they can only access these LLMs which might not have been developed with their needs and priorities in mind. An open and collaborative development approach that is the cornerstone of open-source communities \citep{johnson2006collaboration} is thus neglected in the over-focus on chasing the best single model, underrepresenting the voices and needs of actual LLM users that go beyond synthetic benchmark numbers.

\begin{figure*}
    \centering
    \includegraphics[width=\linewidth]{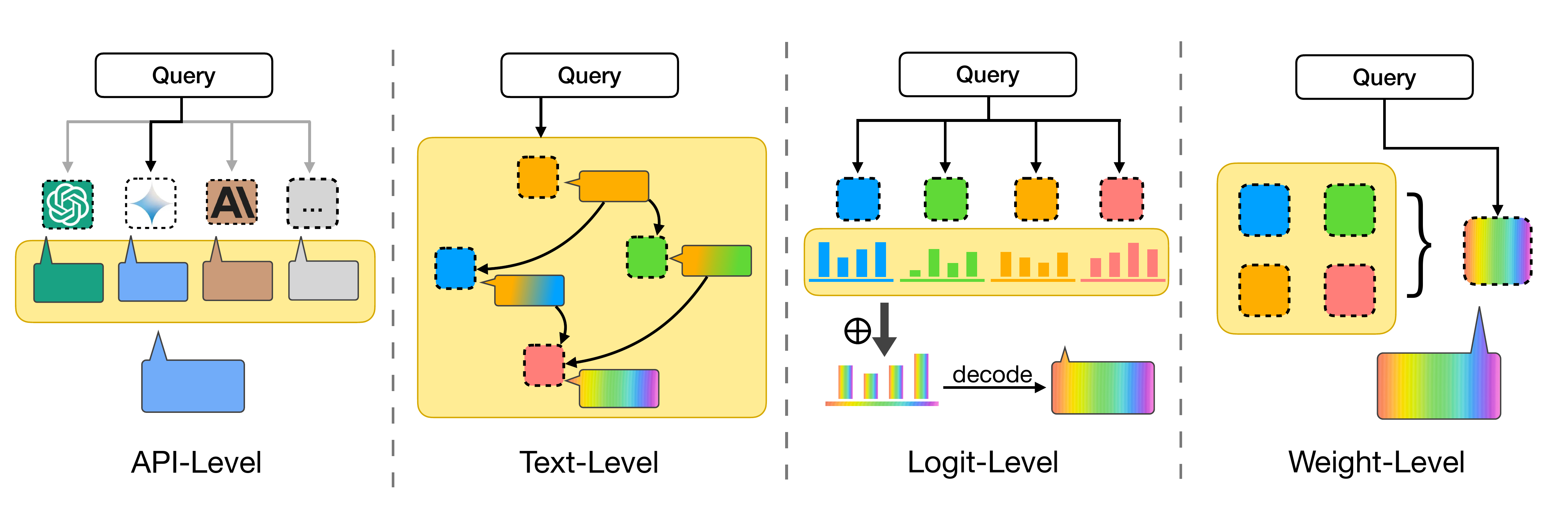}
    \vspace*{-20pt}
    \caption{We propose a typology of multi-LLM collaboration approaches, focusing on different levels of access to LLMs, and survey existing methods that fall into each type.}
    \label{fig:overview}
\end{figure*}

\paragraph{Challenges to Improve One Model's Coverage} 
A tempting solution to these problems of underrepresentation is to further train the current best LLM to improve the representation of data, skills, and users. We argue that this band-aid approach is challenging at best:

When \emph{data} is underrepresented, model developers can scrape from previously unseen domains and perform further fine-tuning. However, it is costly to frequently re-train and update model versions with gigantic LLMs, while private and copyrighted data simply should not be included in LLM training data. Retrieval-augmented geneartion \citep{guu2020retrieval, shi2023replug} could provide unseen data as context, but it is unclear whether LLMs would fully leverage the context \citep{shi-etal-2024-trusting} and to what extent is this steerability reliable \citep{sprague2024cot}.

When \emph{skills} are underrepresented, model developers can derive targeted supervised fine-tuning data for continual learning \citep{zhang2023instruction}. However, tuning to patch a weakness in skills may lead to tradeoffs in other tasks and sometimes even catastrophic forgetting \citep{luo2023empirical, lin2024mitigating}, as any specialization on the trained model might harm its general-purpose utility.

When \emph{humans} are underrepresented, model developers can survey the needs of diverse populations and communities for LLMs and invite collaborative contributions \citep{fengknowledge}. However, there is little to no incentive for teams behind commercial state-of-the-art LLMs to take great strides towards equitable language technologies without obvious profitable gains.

It is important to note that these underrepresentation issues of a single LLM, especially with respect to data and skills, are grounded in \emph{empirical evidence}, i.e., current state-of-the-art LLMs are suffering from these challenges. There might emerge future ``perfect'' algorithms/architectures/etc.~that fully address these issues, but given that multi-LLM collaboration research is \emph{already demonstrating empirical benefits} in addressing these issues, we advocate for multi-LLM collaboration as a promising and effective research avenue. 


\section{Types of Multi-LLM Collaboration}
\label{sec:how}
\vspace{5pt}

We categorize existing (often unrelated) method into a conceptual family of multi-LLM collaboration strategies, organizing the methods by (1) collaboration at different levels of access to an LLM, as illustrated in Figure~\ref{fig:overview}, and (2) collaboration at different stages of LLM's lifecycle: (pre)training, post-training, and inference.

\subsection{Collaboration at different levels of model access}
\paragraph{API-Level} As the name suggests, access to an LLM's  API is sufficient to enable API-level collaboration between models. Such strategies focus on the dynamic selection of the most cost-efficient and high-performing model among a diverse pool of LLMs for different inputs. Intuitively, we should assign simpler requests to smaller \citep{tambon2024assessing}, more efficient LLMs for \emph{reduced cost and latency}, and domain-specific requests to expert LLMs for \emph{improved performance}. There are two mainstream lines of research on API-level LLM collaboration: \emph{Routing}~\citep{hu2024routerbench} and \emph{Cascading}~\citep{chen2023frugalgpt}. 

\emph{Routing} selects the most suitable model only based on the input, without performing inference on any LLM. A typical routing paradigm involves designing a separate router model that learns from human preferences. As a key step, preference labels~\citep{ong2024routellm} that represent the relative response quality of different LLMs are collected for each input.
Prior work developed various router models to learn from input--preference pairs, including non-parametric routers like KNN-based router~\citep{shnitzer2023large}, and parametric routers like MLP-~\citep{hu2024routerbench}, encoder-~\citep{ding2024hybrid}, and decoder-based~\citep{ong2024routellm} routers. Beyond preference data, additional information can be leveraged to assist in routing decision-making. To select the most suitable expert LLM, domain-specific routing strategies~\citet{lu2023routing, stripelis2024tensoropera} extract key information about the task and domain directly from the input. \citet{feng2024graphrouter} further introduce a heterogeneous graph framework to leverage contextual interactions among tasks, queries, and LLMs.

\emph{Cascading} defers the input to larger/more capable LLMs when the response from the smaller LLM is not satisfactory enough. The crux of cascading is the \emph{deferral rule} to determine whether to terminate and return the prediction or to invoke the next LLM. The pioneering work Frugal GPT~\cite{chen2023frugalgpt} trains a regression model that predicts a response quality score and establishes the deferral rule by thresholding the predicted score. \citet{yue2023large} presents a consistency-based approach that estimates the response confidence score, such that inputs with low response confidence are deferred to the next LLM. \citet{gupta2024language} further incorporate token-level uncertainty into deferral rules. Cascading strategies, while potentially improving overall quality by leveraging additional signal from smaller LLMs, often come with increased cost and latency due to the overhead of decoding intermediate responses.

\paragraph{Text-Level} Text-level approaches enable multi-LLM collaboration through exchanges of generated texts, where one LLM's output becomes another LLM's input. They usually follow a conversational setting where LLMs can ``cooperate'' or ``compete'' with each other.

For cooperation, models can \emph{divide and conquer} complex problems through multi-agent systems where each agent is seeded by different models/prompts \citep{wu2024autogen, guo2024large}; specialized models can \emph{augment} general-purpose LLMs to patch their gaps \citep{fengknowledge, shen-etal-2024-learning}; one LLM can generate \emph{feedback} or perform verification for another LLM's outputs and consequently refine the generation \citep{burnsweak, feng-etal-2024-dont}.

For \emph{competition}, multiple LLMs can ``debate'' with each other to advance factuality and reasoning \citep{liang2023encouraging, duimproving}. Recent research also explored employing a pool of diverse specialized LLMs to model social \citep{zhao2024language} and economic \citep{zhaocompeteai} behavior to simulate the real-world environment.

Much effort of multi-LLM collaboration research currently operates at the text-level, probably because such interaction allows for the use of APIs in closed models, the ease of engineering to redirect model outputs, and transparency through intermediate model outputs. However, text-level multi-LLM collaboration also faces challenges such as error propagation from outputs of individual models, the lack of generalization across tasks, and the costs of model inference for multiple LLMs.

\paragraph{Logit-Level}

LLMs may also collaborate by jointly contributing to each next-token prediction. 
In this case, the logit-level predictions of multiple LLMs are combined via arithmetics to create a single next-token logit distribution, which is then normalized into a probability distribution.
This approach uses other LLMs as ``experts'' and/or ``anti-experts'', whose predictions are additively or negatively combined in the prediction, respectively.

Using an anti-expert achieves the effect of steering \textit{away} from the preferences of that model, and is also known as \textit{contrastive decoding} \citep{li-etal-2023-contrastive}.
For instance, the anti-expert may be an LLM tuned explicitly to be toxic \citep{liu-etal-2021-dexperts}, to achieve safer generations, or a smaller LLM \cite{li-etal-2023-contrastive}, to avoid the pitfalls of weaker LMs for better open-ended generation.
In fact, the anti-expert does not need to be a distinct LLM, and can instead be the result of ablating some part of the current LLM, e.g., by withholding necessary context \citep{pei-etal-2023-preadd, sennrich-etal-2024-mitigating, Leng_2024_CVPR, shi-etal-2024-trusting} or early-exiting from an earlier layer of the transformer model \citep{gera-etal-2023-benefits, chuang2024dola}.

On the other hand, using multiple expert LLMs combines their predictions in a product-of-experts fashion.
Intuitively, this leads to next-token predictions that are high-probability under all LLMs.
This has been used to achieve decoding-time adaptation of LLMs using small tuned experts with a large pretrained LLM \cite{liu-etal-2024-tuning, mitchell2024an}, allowing for on-the-fly customization of the weights of multiple finetuning objectives \cite{shi-etal-2024-decoding}.

The weights assigned to experts and anti-experts may also be automatically determined at each time step \cite{mavromatis2024pack, fan2024on, du2024mogu}.
At the extreme, this takes the form of token-level routing among models \citep{shen-etal-2024-learning}.

\paragraph{Weight-Level} 
The collaboration of multiple LLMs through parameter-level collaboration has been explored using paradigms such as mixtures of feed-forward layers \citep{sukhbaatar2024branch}, adapters \citep{wang2022adamixmixtureofadaptationsparameterefficientmodel, pfeiffer2020adapterfusion}, and low-rank adaptation (LoRA) experts \citep{wu2024mixture}. In this paradigm, components like feed-forward layers and adapters are first trained independently on domain-specific or task-specific data to achieve specialization. Subsequently, in a combination stage, these independently trained modules are jointly optimized to collaborate effectively, creating a unified system that benefits from the specialized expertise of each component.
\vspace{5pt}

This framework supports collaboration across varying levels of input granularity the way experts are selected and aggregated. For example, some approaches dynamically select modules for individual \emph{tokens} \citep{vaswani2017attention, houlsby2019parameter, pfeiffer2020adapterfusion, belofsky2023token, wu2024mixture, sukhbaatar2024branch}, enabling fine-grained expertise sharing. 
Others perform collaboration at the \emph{sentence} level \citep{diao2023mixture, xu2023wizardlm}, where different input sentences activate different modules. At the \emph{task} level, methods such as \citet{chiang2024chatbot} assign a single expert model to all examples from a particular task.
Weight-level collaboration typically allows for deeper integration of experts by enabling routing decisions at each layer where modules are inserted, offering greater flexibility and adaptability to diverse tasks and data.
\vspace{5pt}

Another line of weight-level collaboration research is the merging/composition of model weights across multiple LLMs. These approaches mainly vary by \emph{data dependency}, i.e., how much task-specific data is required to compose and adapt models. \emph{Zero-shot} model composition approaches rely on heuristics about model weights \citep{yu2024language, yadav2024ties} or task arithmetic \citep{ilharcoediting} to produce composed models and advance generalization without access to task data. Given a small set of task data, \emph{dynamic composition} approaches optimize the model composition based on performance and metrics on the task data \citep{huang2023lorahub} with perplexity heuristics \citep{mavromatis2024pack} and evolutionary algorithms \citep{feng2024model}. If supervised data is abundant, \emph{learn-to-fuse} approaches design trainable modules \citep{bansalllm}, adapters \citep{wangfusing}, or even LLMs \citep{jiang2023llm} to ``glue'' multiple LLMs together: the component LLMs are often kept frozen while the trainable parts go through supervised fine-tuning from scratch. Weight-level approaches offer a spectrum of solutions based on data availability, and the \emph{many-to-one} setup offers reduced inference costs. However, weight-level approaches are less interpretable in how model expertise is composed and do not tap into the power of collaborative generation like text- or logit-level approaches.

\subsection{Collaboration at different stages of LLM development}
We can also categorize multi-LLM collaboration approaches by the three stages of the LLM lifecycle: \emph{(pre)training, post-training, and inference}. Pretraining-time approaches focus on partitioning LLM training data \citep{gururangan2023scaling} and training multiple specialized LLMs separately \citep{li2022branch} or at the same time \citep{kudugunta2023matformer}. Post-training approaches explore collaborative alignment with modular reward models \citep{jang2023personalized}, multi-LLM self-alignment \citep{feng2024modular}, or constructing synthetic supervised fine-tuning data through multi-LLM debate \citep{subramaniam2024debategpt, subramaniam2025multiagent}. The vast majority of multi-LLM collaboration approaches currently operate at inference time, offering diverse ways of reusing existing models spanning all four collaboration levels \citep{hu2024routerbench, duimproving, liu-etal-2024-tuning}. In general, weight-level methods often require more (pre)training and post-training efforts, while API-level/logit-level collaborations focus more on inference-time solutions.

By conceptually structuring and organizing these (originally unrelated) methods into a family of approaches, we argue that multi-LLM collaboration research offers flexible methodologies for any level of model access across all stages in the LLM lifecycle, providing an alternative and promising school of thought to advance language technologies.

\section{The Promise of Multi-LLM Collaboration}
\label{sec:what}
\vspace{5pt}

Multi-LLM collaboration systems offer unique advantages over single general-purpose models: we summarize the methodological and empirical benefits of existing multi-LLM proposals in this section.

\paragraph{Factuality and reliability} Despite prior efforts \citep{shi2023replug, Press2022MeasuringAN, feng2023cook} to expand the knowledge of LLMs, knowledge gaps---missing or outdated information in LLMs---may persist due to the ever-evolving nature of knowledge, presenting challenges to the reliability of LLM responses. Self-reflection \citep{wang2022self, xu2024sayself, shinn2024reflexion, madaan2024self}, where a single LLM critically evaluates its own generations, is used in decoding, confidence calibration, and inference to improve factual accuracy and mitigate hallucinations. However, this method suffers from confirmation biases \citep{ji2023survey} and relies on the assumption that LLMs can generate novel reflections from their initial outputs \citep{liang2023encouraging}. Recent studies address these issues by promoting collaboration between multiple LLMs. With distinct knowledge gaps, LLMs evaluate and reflect on each other's outputs, collaboratively probing and identifying the knowledge gaps of each other. Specifically, \citet{feng2024don} enable robust LLM abstention through multi-LLM collaboration to reflect on generated text in cooperative or competitive settings. \citet{cohen2023lm} employ cross-examination to detect errors in LLM generations. Other studies \citep{xiong2023examining, liang2023encouraging, duimproving} suggest that multiple LLMs could propose and debate their individual responses and reasoning processes over multiple rounds to arrive at a common final answer, and LLMs with comparable abilities have been shown to demonstrate such collaborative spirit \citep{xiong2023examining}. Given its superior performance in various experiment settings, we believe that multi-LLM collaboration offers a promising way to further improve the factual validity of generated context and reduce fallacious answers and hallucinations that contemporary models are prone to.

\paragraph{Alignment and pluralism} State-of-the-art LLMs are documented with all kinds of cultural \citep{naous2023having}, political \citep{santurkar2023whose}, and broadly social biases \citep{kumar2023language}. This comes with the fact that these models have already seen ``diverse'' web data that should serve as a decentralized representation of real-world diversity. Much research attributes this to LLMs learning disproportionately from and hence reinforcing the majority in training data \citep{feng2023pretraining, gallegos2024bias}, thus scaling data diversity used in training a single LLM is not an effective solution. We see an increasing line of work focused on \emph{modular multi-LLM systems} to alleviate these biases, including modular plug-ins \citep{feng2024modular}, multi-LLM as a judge \citep{zhao2024language}, and employing multiple and compositional reward models \citep{jang2023personalized}. Together with data-side modularity spanning diverse communities \citep{kumar2024compo, kirk2024prism} we believe multi-LLM collaboration offers a modular and flexible solution to addressing the fairness and pluralism challenges of LLMs.

\paragraph{Efficiency} The most capable LLMs at the moment often feature gargantuan sizes and prohibitively high inference costs. However, not all queries require such computation overhead: by employing multi-LLM collaboration across sizes/expertise the largest model doesn't need to be called every single time. MatFormer \citep{kudugunta2023matformer} simultaneously trains modules of varying sizes in a nested architecture and could be selectively activated to result in LLMs of varying sizes given the compute budget. Instead of training an LLM on \emph{all} the data, approaches such as Branch-Train-Merge \citep{li2022branch} leverage the modularity of data provenance to train a pool of LLM experts and dynamically aggregated for inference. A growing line of research also investigates \emph{defer} and \emph{backoff} behavior between models of varying sizes and/or specialization \citep{shen-etal-2024-learning, jung2024trust}. These approaches highlight multi-LLM collaboration as a promising direction to balance utility and training/inference efficiency.

\paragraph{Adaptation} Training a gigantic LLM and re-purposing it with prompt engineering is the most popular status quo of LLM research and applications. However, one gigantic model is prohibitively expensive to re-train and update, while the effectiveness of prompt-based adaptation is limited and brittle \citep{sprague2024cot}. Multi-LLM collaboration offers strategies for adapting language models that are lightweight and flexible: Token-level methods pair a general-purpose LLM with specialized peers for collaborative generation \citep{shen-etal-2024-learning}; logit-level approaches mixes the logit distributions of multiple LLMs for collaborative decoding \citep{liu-etal-2024-tuning}; weight-level approaches flexibly reuse and adapt existing models/adapters through weight arithmetic \citep{ilharcoediting, han2023ssd, yadav2024ties, feng2024model}. Multi-LLM collaboration offers diverse and flexible solutions for adaptation spanning varying levels of model access.

\paragraph{Privacy} Despite the extensive effort to curate (pre)training data, private and copyrighted data will need to be left out for privacy, compliance, and ethics concerns \citep{karamolegkou2023copyright, yao2024survey}. These data sets are nonetheless helpful in highly specialized or personalized contexts. Multi-LLM offers preliminary solutions where private/copyrighted data could be employed in a local model at the data provenance, then interact with a larger general-purpose model \citep{zhang2024cogenesis}. Though it might be possible to extract private data from the model \citep{carlini2021extracting}, we envision future work on augmenting the ``private'' LLM with contextual integrity guardrails \citep{mireshghallahcan} for controllable and context-aware information sharing.

\paragraph{Democratization and collaborative development} A single LLM is often trained by only a team of researchers and engineers, struggling to reflect the diversity of real-world LLM users. The priorities of long-tail and underprivileged users are often not incorporated when making decisions about model training and alignment. On the contrary, multi-LLM collaboration uniquely enables decentralized and collaborative development: all stakeholders in LLM development and applications could contribute models based on their needs, priorities, and compute budgets, then composed through various levels of multi-LLM collaboration protocols (\Sref{sec:how}). In this way, we democratize language technologies through participatory and collaborative development where everyone is welcome.

% ...


\section{Future Directions for Multi-LLM Collaboration Research}
\label{sec:discussion}

We identify various limitations of existing multi-LLM collaboration systems and motivate future work.

\paragraph{Theories of human communication} While the current approach focuses on developing a single general-purpose LLM, there is no ``general-purpose'' human, but specialized individuals collaborating through various communication protocols for collective intelligence \citep{hutchins2000distributed}. We thus argue that future multi-LLM collaboration research could benefit from cognitive science and communications theories, designing social science-inspired protocols for multiple LLMs to compose and collaborate.

\paragraph{Encapsulation and handoff} Another interesting challenge in multi-LLM collaboration is the absence of clear handoff boundaries. In software engineering, \emph{encapsulation} serves as a cornerstone of collaborative development by establishing well-defined interfaces between components: modifications to one part of the codebase do not propagate unexpected changes to others. However, especially in weight-level LLM collaboration, cleanly separating and containing the expertise of different models remains an open challenge. While recent work has demonstrated progress in developing modularized model components~\citep{pfeiffer2020adapterfusion, hu2021lora,yadav2024survey}, modifications to base model weights can still introduce unpredictable behavioral changes beyond the intended training objectives (for example, catastrophic forgetting~\citep{mccloskey1989catastrophic, kirkpatrick2017overcoming}). Developing reliable encapsulation mechanisms can ensure robust and predictable model composition, and could be a critical step to achieve the vision for ``building LMs like open-source software''~\citep{raffel2021call}.

\paragraph{Compatibility with the status quo} Despite the active research in multi-LLM collaboration, there is limited uptake in large-scale and industry settings beyond academic papers. One reason could be that many existing multi-LLM approaches require the training/development of extra modules such as gates and routers \citep{jiang2023llm, muqeethlearning}, while most open-source activities only feature the sharing of model weights. We thus argue that future multi-LLM protocols should be compatible with the status quo of model sharing by employing limited to no extra step beyond employing existing model checkpoints.

\paragraph{Interpretability insights} 
Interpretability techniques unveil the mechanisms underlying language models for reasoning \citep{stolfo-etal-2023-mechanistic}, factual association \citep{meng2022locating}, and more \citep{nandaprogress}. The interpretability insights enable localized manipulation of sub-modules for efficient enhancement and editing \citep{yin2024lofit}, thereby facilitating the potential for lightweight multi-model collaboration.
Moreover, while diverse language models may exhibit similar or distinct mechanisms for comparable tasks, the reliability of their capability beyond mere memorization varies \citep{yang-etal-2024-large-language-models}. Interpretability tools offer insights into determining the fitting weight/contribution of each component model in multi-LLM collaboration and could lead to improved collaboration outcomes.

\paragraph{Evaluating multi-LLM collaboration} Research on modular and multi-LLM systems has not yet devised an agreed-upon and detailed evaluation methodology. Most of the existing work resorts to evaluation with tasks and datasets typical for a single LLM. Future work could explore specifically evaluating multi-LLM collaboration, designing tasks and datasets where multiple models are evaluated separately and in collaboration, e.g., ablating by withholding copyright data \citep{minsilo}, or evaluating multi-agent collaboration where multiple models divide and conquer complex problems \citep{guo2024large}.

\paragraph{Democratizing ways of contribution} While we hope that collaborative and participatory contributions to multi-LLM systems could alleviate the underrepresentation of people, \emph{not everyone knows how to train an LLM and contribute}. This is especially true for the already underrepresented and underprivileged \citep{kirk2024prism}, thus the benefits of multi-LLM collaboration will not reach them if we expect users to train and contribute models on their own. Thus, we argue that we should lower the barrier of contribution: for example, by designing an agent framework that automatically solicits user feedback in natural language, fetches data, trains models, generates synthetic data to evaluate, and finally pushes the model and contribute. In this way, users only need to provide a few sentences of feedback about the gaps in existing LLMs, and a new component LLM could be developed and contributed on their behalf.

\section{Alternative Views}

We identify two alternative views to our position.

\emph{We could patch the underrepresentations of data, skills, and people by further augmenting a single model.} While existing band-aid approaches such as LoRA fine-tuning \citep{hu2021lora} or retrieval augmented generation (RAG) \citep{shi2023replug, jiang2023active, asaiself} patch the gaps in data and skills to some extent, we present empirical evidence of their limitations in Section \ref{sec:why}, suffering from challenges such as privacy and copyright, catastrophic forgetting, lack of participation, and more. Further fine-tuning with LoRA could patch the gap of skills, but it risks jeopardizing the general-purposeness and leads to tradeoffs of existing skills \citep{kirkpatrick2017overcoming}; retrieval could provide new information and data to improve reliability, but there is no guarantee that LLMs would fully leverage the retrieved context \citep{shi-etal-2024-trusting}. While it is not impossible that with future progress a single LLM could offer perfect representations, we argue that multi-LLM collaboration offers a more concrete and actionable roadmap to advance language technologies, and a more efficient one, as it reuses developments made so far.

\emph{We could enable collaboration through a single model.} It is theoretically possible to collaborate in the development lifecycle of a single model. Different communities could contribute heterogeneous data to be combined for training a single model; different engineering teams could train part of the model architecture for later merging; different users could annotate diverse alignment preferences to jointly align an LLM. We argue that while they are all possible, it is more natural to collaborate on the level of models since 1) model sharing is the default open-source activity, 2) there are already 1,261,059 LLMs\footnote{Huggingface accessed on Jan 6, 2025.} openly available for collaboration, and 3) the companies that have the resource to carry out these protocols are incentivized to not go open about development of their LLM for competitive advantage. We envision multi-LLM collaboration as a promising path to reuse existing models, promote collaborative development, and advance compositional intelligence.


\section{Related Work}

Two recent works discuss related topics. 

\citet{yadav2024survey} present a taxonomy of model merging and mixture-of-experts (MoE) approaches, arguing for reusing and routing of existing expert models. They focus primarily on weight-level collaboration approaches, while we aggregate a broader family of methods with a broader definition of \emph{multi-LLM collaboration} where models could collaborate through four different levels of information exchange.

\citet{duposition} present a position paper arguing for compositional generative modeling, discussing the benefits of combining multiple modules across computer vision, reinforcement learning, robotics, and a brief mention of language. We specifically focus on language models and dive deep into LLM-specific arguments, methods, and future research.

\section{Conclusion}

We argue that one LLM is not enough and advocate for multi-LLM collaboration to better represent diverse data distributions, heterogeneous skills, and pluralistic populations. We propose a hierarchy of existing multi-LLM collaboration approaches based on information exchange levels, spanning API-level, text-level, logit-level, and weight-level collaboration. We then summarize the benefits of existing multi-LLM systems over a single model and discuss the limitations of existing methods to motivate future work. We envision multi-LLM collaboration as a viable path to compositional intelligence and an important initiative toward collaborative AI development.

\bibliography{example_paper}

\begin{thebibliography}{133}
\providecommand{\natexlab}[1]{#1}
\providecommand{\url}[1]{\texttt{#1}}
\expandafter\ifx\csname urlstyle\endcsname\relax
  \providecommand{\doi}[1]{doi: #1}\else
  \providecommand{\doi}{doi: \begingroup \urlstyle{rm}\Url}\fi

\bibitem[Achiam et~al.(2023)Achiam, Adler, Agarwal, Ahmad, Akkaya, Aleman, Almeida, Altenschmidt, Altman, Anadkat, et~al.]{achiam2023gpt}
Achiam, J., Adler, S., Agarwal, S., Ahmad, L., Akkaya, I., Aleman, F.~L., Almeida, D., Altenschmidt, J., Altman, S., Anadkat, S., et~al.
\newblock Gpt-4 technical report.
\newblock \emph{arXiv preprint arXiv:2303.08774}, 2023.

\bibitem[Asai et~al.(2024)Asai, Wu, Wang, Sil, and Hajishirzi]{asaiself}
Asai, A., Wu, Z., Wang, Y., Sil, A., and Hajishirzi, H.
\newblock Self-rag: Learning to retrieve, generate, and critique through self-reflection.
\newblock In \emph{The Twelfth International Conference on Learning Representations}, 2024.

\bibitem[Bansal et~al.(2024)Bansal, Samanta, Dalmia, Gupta, Ganapathy, Bapna, Jain, and Talukdar]{bansalllm}
Bansal, R., Samanta, B., Dalmia, S., Gupta, N., Ganapathy, S., Bapna, A., Jain, P., and Talukdar, P.
\newblock Llm augmented llms: Expanding capabilities through composition.
\newblock In \emph{The Twelfth International Conference on Learning Representations}, 2024.

\bibitem[Belofsky(2023)]{belofsky2023token}
Belofsky, J.
\newblock Token-level adaptation of lora adapters for downstream task generalization.
\newblock In \emph{Proceedings of the 2023 6th Artificial Intelligence and Cloud Computing Conference}, pp.\  168--172, 2023.

\bibitem[Brown et~al.(2020)Brown, Mann, Ryder, Subbiah, Kaplan, Dhariwal, Neelakantan, Shyam, Sastry, Askell, Agarwal, Herbert-Voss, Krueger, Henighan, Child, Ramesh, Ziegler, Wu, Winter, Hesse, Chen, Sigler, teusz Litwin, Gray, Chess, Clark, Berner, McCandlish, Radford, Sutskever, and Amodei]{Brown2020LanguageMA}
Brown, T.~B., Mann, B., Ryder, N., Subbiah, M., Kaplan, J., Dhariwal, P., Neelakantan, A., Shyam, P., Sastry, G., Askell, A., Agarwal, S., Herbert-Voss, A., Krueger, G., Henighan, T., Child, R., Ramesh, A., Ziegler, D.~M., Wu, J., Winter, C., Hesse, C., Chen, M., Sigler, E., teusz Litwin, M., Gray, S., Chess, B., Clark, J., Berner, C., McCandlish, S., Radford, A., Sutskever, I., and Amodei, D.
\newblock Language models are few-shot learners.
\newblock \emph{ArXiv}, abs/2005.14165, 2020.

\bibitem[Burns et~al.(2024)Burns, Izmailov, Kirchner, Baker, Gao, Aschenbrenner, Chen, Ecoffet, Joglekar, Leike, et~al.]{burnsweak}
Burns, C., Izmailov, P., Kirchner, J.~H., Baker, B., Gao, L., Aschenbrenner, L., Chen, Y., Ecoffet, A., Joglekar, M., Leike, J., et~al.
\newblock Weak-to-strong generalization: Eliciting strong capabilities with weak supervision.
\newblock In \emph{Forty-first International Conference on Machine Learning}, 2024.

\bibitem[Carlini et~al.(2021)Carlini, Tramer, Wallace, Jagielski, Herbert-Voss, Lee, Roberts, Brown, Song, Erlingsson, et~al.]{carlini2021extracting}
Carlini, N., Tramer, F., Wallace, E., Jagielski, M., Herbert-Voss, A., Lee, K., Roberts, A., Brown, T., Song, D., Erlingsson, U., et~al.
\newblock Extracting training data from large language models.
\newblock In \emph{30th USENIX Security Symposium (USENIX Security 21)}, pp.\  2633--2650, 2021.

\bibitem[Chen et~al.(2023)Chen, Zaharia, and Zou]{chen2023frugalgpt}
Chen, L., Zaharia, M., and Zou, J.
\newblock Frugalgpt: How to use large language models while reducing cost and improving performance.
\newblock \emph{arXiv preprint arXiv:2305.05176}, 2023.

\bibitem[Chen et~al.(2024)Chen, Asai, Mireshghallah, Min, Grimmelmann, Choi, Hajishirzi, Zettlemoyer, and Koh]{chen2024copybench}
Chen, T., Asai, A., Mireshghallah, N., Min, S., Grimmelmann, J., Choi, Y., Hajishirzi, H., Zettlemoyer, L., and Koh, P.~W.
\newblock Copybench: Measuring literal and non-literal reproduction of copyright-protected text in language model generation.
\newblock In \emph{Proceedings of the 2024 Conference on Empirical Methods in Natural Language Processing}, pp.\  15134--15158, 2024.

\bibitem[Chiang et~al.(2024{\natexlab{a}})Chiang, Zheng, Sheng, Angelopoulos, Li, Li, Zhang, Zhu, Jordan, Gonzalez, et~al.]{chiang2024chatbot}
Chiang, W.-L., Zheng, L., Sheng, Y., Angelopoulos, A.~N., Li, T., Li, D., Zhang, H., Zhu, B., Jordan, M., Gonzalez, J.~E., et~al.
\newblock Chatbot arena: An open platform for evaluating llms by human preference.
\newblock \emph{arXiv preprint arXiv:2403.04132}, 2024{\natexlab{a}}.

\bibitem[Chiang et~al.(2024{\natexlab{b}})Chiang, Zheng, Sheng, Angelopoulos, Li, Li, Zhu, Zhang, Jordan, Gonzalez, et~al.]{chiangchatbot}
Chiang, W.-L., Zheng, L., Sheng, Y., Angelopoulos, A.~N., Li, T., Li, D., Zhu, B., Zhang, H., Jordan, M., Gonzalez, J.~E., et~al.
\newblock Chatbot arena: An open platform for evaluating llms by human preference.
\newblock In \emph{Forty-first International Conference on Machine Learning}, 2024{\natexlab{b}}.

\bibitem[Chuang et~al.(2024)Chuang, Xie, Luo, Kim, Glass, and He]{chuang2024dola}
Chuang, Y.-S., Xie, Y., Luo, H., Kim, Y., Glass, J.~R., and He, P.
\newblock Dola: Decoding by contrasting layers improves factuality in large language models.
\newblock In \emph{The Twelfth International Conference on Learning Representations}, 2024.

\bibitem[Cohen et~al.(2023)Cohen, Hamri, Geva, and Globerson]{cohen2023lm}
Cohen, R., Hamri, M., Geva, M., and Globerson, A.
\newblock Lm vs lm: Detecting factual errors via cross examination.
\newblock \emph{arXiv preprint arXiv:2305.13281}, 2023.

\bibitem[Devvrit et~al.(2024)Devvrit, Kudugunta, Kusupati, Dettmers, Chen, Dhillon, Tsvetkov, Hajishirzi, Kakade, Farhadi, , and Jain]{kudugunta2023matformer}
Devvrit, F., Kudugunta, S., Kusupati, A., Dettmers, T., Chen, K., Dhillon, I.~S., Tsvetkov, Y., Hajishirzi, H., Kakade, S.~M., Farhadi, A., , and Jain, P.
\newblock Matformer: Nested transformer for elastic inference.
\newblock In \emph{NeurIPS}, 2024.

\bibitem[Dhingra et~al.(2022)Dhingra, Cole, Eisenschlos, Gillick, Eisenstein, and Cohen]{dhingra2022time}
Dhingra, B., Cole, J.~R., Eisenschlos, J.~M., Gillick, D., Eisenstein, J., and Cohen, W.~W.
\newblock Time-aware language models as temporal knowledge bases.
\newblock \emph{Transactions of the Association for Computational Linguistics}, 10:\penalty0 257--273, 2022.

\bibitem[Diao et~al.(2023)Diao, Xu, Xu, Wang, and Zhang]{diao2023mixture}
Diao, S., Xu, T., Xu, R., Wang, J., and Zhang, T.
\newblock Mixture-of-domain-adapters: Decoupling and injecting domain knowledge to pre-trained language models memories.
\newblock \emph{arXiv preprint arXiv:2306.05406}, 2023.

\bibitem[Ding et~al.(2024)Ding, Mallick, Wang, Sim, Mukherjee, Ruhle, Lakshmanan, and Awadallah]{ding2024hybrid}
Ding, D., Mallick, A., Wang, C., Sim, R., Mukherjee, S., Ruhle, V., Lakshmanan, L.~V., and Awadallah, A.~H.
\newblock Hybrid llm: Cost-efficient and quality-aware query routing.
\newblock \emph{arXiv preprint arXiv:2404.14618}, 2024.

\bibitem[Du \& Kaelbling(2024)Du and Kaelbling]{duposition}
Du, Y. and Kaelbling, L.~P.
\newblock Position: Compositional generative modeling: A single model is not all you need.
\newblock In \emph{Forty-first International Conference on Machine Learning}, 2024.

\bibitem[Du et~al.(2024{\natexlab{a}})Du, Li, Torralba, Tenenbaum, and Mordatch]{duimproving}
Du, Y., Li, S., Torralba, A., Tenenbaum, J.~B., and Mordatch, I.
\newblock Improving factuality and reasoning in language models through multiagent debate.
\newblock In \emph{Forty-first International Conference on Machine Learning}, 2024{\natexlab{a}}.

\bibitem[Du et~al.(2024{\natexlab{b}})Du, Zhao, Zhao, Ma, Chen, Huo, Yang, Xu, and Qin]{du2024mogu}
Du, Y., Zhao, S., Zhao, D., Ma, M., Chen, Y., Huo, L., Yang, Q., Xu, D., and Qin, B.
\newblock Mo{GU}: A framework for enhancing safety of {LLM}s while preserving their usability.
\newblock In \emph{The Thirty-eighth Annual Conference on Neural Information Processing Systems}, 2024{\natexlab{b}}.

\bibitem[EEOC(2024)]{workforce}
EEOC.
\newblock High tech, low inclusion: Diversity in the high tech workforce and sector 2014 - 2022, 2024.

\bibitem[Faisal et~al.(2024)Faisal, Ahia, Srivastava, Ahuja, Chiang, Tsvetkov, and Anastasopoulos]{faisal-etal-2024-dialectbench}
Faisal, F., Ahia, O., Srivastava, A., Ahuja, K., Chiang, D., Tsvetkov, Y., and Anastasopoulos, A.
\newblock {DIALECTBENCH}: An {NLP} benchmark for dialects, varieties, and closely-related languages.
\newblock In Ku, L.-W., Martins, A., and Srikumar, V. (eds.), \emph{Proceedings of the 62nd Annual Meeting of the Association for Computational Linguistics (Volume 1: Long Papers)}, 2024.

\bibitem[Fan et~al.(2024)Fan, Lu, Wei, Tian, Qu, Chen, and Cheng]{fan2024on}
Fan, C., Lu, Z., Wei, W., Tian, J., Qu, X., Chen, D., and Cheng, Y.
\newblock On giant's shoulders: Effortless weak to strong by dynamic logits fusion.
\newblock In \emph{The Thirty-eighth Annual Conference on Neural Information Processing Systems}, 2024.

\bibitem[Feng et~al.(2023{\natexlab{a}})Feng, Park, Liu, and Tsvetkov]{feng2023pretraining}
Feng, S., Park, C.~Y., Liu, Y., and Tsvetkov, Y.
\newblock From pretraining data to language models to downstream tasks: Tracking the trails of political biases leading to unfair nlp models.
\newblock In \emph{Proceedings of the 61st Annual Meeting of the Association for Computational Linguistics (Volume 1: Long Papers)}, pp.\  11737--11762, 2023{\natexlab{a}}.

\bibitem[Feng et~al.(2023{\natexlab{b}})Feng, Shi, Bai, Balachandran, He, and Tsvetkov]{feng2023cook}
Feng, S., Shi, W., Bai, Y., Balachandran, V., He, T., and Tsvetkov, Y.
\newblock Cook: Empowering general-purpose language models with modular and collaborative knowledge.
\newblock \emph{arXiv preprint arXiv:2305.09955}, 2023{\natexlab{b}}.

\bibitem[Feng et~al.(2024{\natexlab{a}})Feng, Shi, Bai, Balachandran, He, and Tsvetkov]{fengknowledge}
Feng, S., Shi, W., Bai, Y., Balachandran, V., He, T., and Tsvetkov, Y.
\newblock Knowledge card: Filling llms' knowledge gaps with plug-in specialized language models.
\newblock In \emph{The Twelfth International Conference on Learning Representations}, 2024{\natexlab{a}}.

\bibitem[Feng et~al.(2024{\natexlab{b}})Feng, Shi, Wang, Ding, Balachandran, and Tsvetkov]{feng-etal-2024-dont}
Feng, S., Shi, W., Wang, Y., Ding, W., Balachandran, V., and Tsvetkov, Y.
\newblock Don't hallucinate, abstain: Identifying {LLM} knowledge gaps via multi-{LLM} collaboration.
\newblock In \emph{Proceedings of the 62nd Annual Meeting of the Association for Computational Linguistics (Volume 1: Long Papers)}, August 2024{\natexlab{b}}.

\bibitem[Feng et~al.(2024{\natexlab{c}})Feng, Shi, Wang, Ding, Balachandran, and Tsvetkov]{feng2024don}
Feng, S., Shi, W., Wang, Y., Ding, W., Balachandran, V., and Tsvetkov, Y.
\newblock Don't hallucinate, abstain: Identifying llm knowledge gaps via multi-llm collaboration.
\newblock \emph{arXiv preprint arXiv:2402.00367}, 2024{\natexlab{c}}.

\bibitem[Feng et~al.(2024{\natexlab{d}})Feng, Sorensen, Liu, Fisher, Park, Choi, and Tsvetkov]{feng2024modular}
Feng, S., Sorensen, T., Liu, Y., Fisher, J., Park, C.~Y., Choi, Y., and Tsvetkov, Y.
\newblock Modular pluralism: Pluralistic alignment via multi-llm collaboration.
\newblock In \emph{Proceedings of the 2024 Conference on Empirical Methods in Natural Language Processing}, pp.\  4151--4171, 2024{\natexlab{d}}.

\bibitem[Feng et~al.(2024{\natexlab{e}})Feng, Wang, Wang, Ebrahimi, Palangi, Miculicich, Kulshrestha, Rauschmayr, Choi, Tsvetkov, et~al.]{feng2024model}
Feng, S., Wang, Z., Wang, Y., Ebrahimi, S., Palangi, H., Miculicich, L., Kulshrestha, A., Rauschmayr, N., Choi, Y., Tsvetkov, Y., et~al.
\newblock Model swarms: Collaborative search to adapt llm experts via swarm intelligence.
\newblock \emph{arXiv preprint arXiv:2410.11163}, 2024{\natexlab{e}}.

\bibitem[Feng et~al.(2024{\natexlab{f}})Feng, Shen, and You]{feng2024graphrouter}
Feng, T., Shen, Y., and You, J.
\newblock Graphrouter: A graph-based router for llm selections.
\newblock \emph{arXiv preprint arXiv:2410.03834}, 2024{\natexlab{f}}.

\bibitem[Fourrier et~al.(2024)Fourrier, Habib, Lozovskaya, Szafer, and Wolf]{open-llm-leaderboard-v2}
Fourrier, C., Habib, N., Lozovskaya, A., Szafer, K., and Wolf, T.
\newblock Open llm leaderboard v2, 2024.

\bibitem[Gallegos et~al.(2024)Gallegos, Rossi, Barrow, Tanjim, Kim, Dernoncourt, Yu, Zhang, and Ahmed]{gallegos2024bias}
Gallegos, I.~O., Rossi, R.~A., Barrow, J., Tanjim, M.~M., Kim, S., Dernoncourt, F., Yu, T., Zhang, R., and Ahmed, N.~K.
\newblock Bias and fairness in large language models: A survey.
\newblock \emph{Computational Linguistics}, pp.\  1--79, 2024.

\bibitem[Gera et~al.(2023)Gera, Friedman, Arviv, Gunasekara, Sznajder, Slonim, and Shnarch]{gera-etal-2023-benefits}
Gera, A., Friedman, R., Arviv, O., Gunasekara, C., Sznajder, B., Slonim, N., and Shnarch, E.
\newblock The benefits of bad advice: Autocontrastive decoding across model layers.
\newblock In \emph{Proceedings of the 61st Annual Meeting of the Association for Computational Linguistics (Volume 1: Long Papers)}, 2023.

\bibitem[Guo et~al.(2025)Guo, Yang, Zhang, Song, Zhang, Xu, Zhu, Ma, Wang, Bi, et~al.]{guo2025deepseek}
Guo, D., Yang, D., Zhang, H., Song, J., Zhang, R., Xu, R., Zhu, Q., Ma, S., Wang, P., Bi, X., et~al.
\newblock Deepseek-r1: Incentivizing reasoning capability in llms via reinforcement learning.
\newblock \emph{arXiv preprint arXiv:2501.12948}, 2025.

\bibitem[Guo et~al.(2024)Guo, Chen, Wang, Chang, Pei, Chawla, Wiest, and Zhang]{guo2024large}
Guo, T., Chen, X., Wang, Y., Chang, R., Pei, S., Chawla, N.~V., Wiest, O., and Zhang, X.
\newblock Large language model based multi-agents: A survey of progress and challenges.
\newblock \emph{arXiv preprint arXiv:2402.01680}, 2024.

\bibitem[Gupta et~al.(2024)Gupta, Narasimhan, Jitkrittum, Rawat, Menon, and Kumar]{gupta2024language}
Gupta, N., Narasimhan, H., Jitkrittum, W., Rawat, A.~S., Menon, A.~K., and Kumar, S.
\newblock Language model cascades: Token-level uncertainty and beyond.
\newblock \emph{arXiv preprint arXiv:2404.10136}, 2024.

\bibitem[Gururangan et~al.(2023)Gururangan, Li, Lewis, Shi, Althoff, Smith, and Zettlemoyer]{gururangan2023scaling}
Gururangan, S., Li, M., Lewis, M., Shi, W., Althoff, T., Smith, N.~A., and Zettlemoyer, L.
\newblock Scaling expert language models with unsupervised domain discovery.
\newblock \emph{arXiv preprint arXiv:2303.14177}, 2023.

\bibitem[Guu et~al.(2020)Guu, Lee, Tung, Pasupat, and Chang]{guu2020retrieval}
Guu, K., Lee, K., Tung, Z., Pasupat, P., and Chang, M.
\newblock Retrieval augmented language model pre-training.
\newblock In \emph{International conference on machine learning}, pp.\  3929--3938. PMLR, 2020.

\bibitem[Han et~al.(2023)Han, Kumar, Tsvetkov, and Ghazvininejad]{han2023ssd}
Han, X., Kumar, S., Tsvetkov, Y., and Ghazvininejad, M.
\newblock Ssd-2: Scaling and inference-time fusion of diffusion language models.
\newblock \emph{arXiv preprint arXiv:2305.14771}, 2023.

\bibitem[Henshall(2024)]{bigtech}
Henshall, W.
\newblock Big tech companies were investors in smaller ai labs. now they’re rivals.
\newblock https://time.com/6977424/ai-competition-openai-anthropic-microsoft-amazon/, 2024.

\bibitem[Hoffmann et~al.(2022)Hoffmann, Borgeaud, Mensch, Buchatskaya, Cai, Rutherford, de~Las~Casas, Hendricks, Welbl, Clark, et~al.]{hoffmann2022training}
Hoffmann, J., Borgeaud, S., Mensch, A., Buchatskaya, E., Cai, T., Rutherford, E., de~Las~Casas, D., Hendricks, L.~A., Welbl, J., Clark, A., et~al.
\newblock Training compute-optimal large language models.
\newblock In \emph{Proceedings of the 36th International Conference on Neural Information Processing Systems}, pp.\  30016--30030, 2022.

\bibitem[Houlsby et~al.(2019)Houlsby, Giurgiu, Jastrzebski, Morrone, De~Laroussilhe, Gesmundo, Attariyan, and Gelly]{houlsby2019parameter}
Houlsby, N., Giurgiu, A., Jastrzebski, S., Morrone, B., De~Laroussilhe, Q., Gesmundo, A., Attariyan, M., and Gelly, S.
\newblock Parameter-efficient transfer learning for nlp.
\newblock In \emph{International conference on machine learning}, pp.\  2790--2799. PMLR, 2019.

\bibitem[Hu et~al.(2021)Hu, Shen, Wallis, Allen-Zhu, Li, Wang, Wang, and Chen]{hu2021lora}
Hu, E.~J., Shen, Y., Wallis, P., Allen-Zhu, Z., Li, Y., Wang, S., Wang, L., and Chen, W.
\newblock Lora: Low-rank adaptation of large language models.
\newblock \emph{arXiv preprint arXiv:2106.09685}, 2021.

\bibitem[Hu et~al.(2024)Hu, Bieker, Li, Jiang, Keigwin, Ranganath, Keutzer, and Upadhyay]{hu2024routerbench}
Hu, Q.~J., Bieker, J., Li, X., Jiang, N., Keigwin, B., Ranganath, G., Keutzer, K., and Upadhyay, S.~K.
\newblock Routerbench: A benchmark for multi-llm routing system.
\newblock \emph{arXiv preprint arXiv:2403.12031}, 2024.

\bibitem[Huang et~al.(2023)Huang, Liu, Lin, Pang, Du, and Lin]{huang2023lorahub}
Huang, C., Liu, Q., Lin, B.~Y., Pang, T., Du, C., and Lin, M.
\newblock Lorahub: Efficient cross-task generalization via dynamic lora composition.
\newblock \emph{arXiv preprint arXiv:2307.13269}, 2023.

\bibitem[Hutchins(2000)]{hutchins2000distributed}
Hutchins, E.
\newblock Distributed cognition.
\newblock \emph{International Encyclopedia of the Social and Behavioral Sciences. Elsevier Science}, 138:\penalty0 1--10, 2000.

\bibitem[Ilharco et~al.(2023)Ilharco, Ribeiro, Wortsman, Schmidt, Hajishirzi, and Farhadi]{ilharcoediting}
Ilharco, G., Ribeiro, M.~T., Wortsman, M., Schmidt, L., Hajishirzi, H., and Farhadi, A.
\newblock Editing models with task arithmetic.
\newblock In \emph{The Eleventh International Conference on Learning Representations}, 2023.

\bibitem[Jang et~al.(2023)Jang, Kim, Lin, Wang, Hessel, Zettlemoyer, Hajishirzi, Choi, and Ammanabrolu]{jang2023personalized}
Jang, J., Kim, S., Lin, B.~Y., Wang, Y., Hessel, J., Zettlemoyer, L., Hajishirzi, H., Choi, Y., and Ammanabrolu, P.
\newblock Personalized soups: Personalized large language model alignment via post-hoc parameter merging.
\newblock \emph{arXiv preprint arXiv:2310.11564}, 2023.

\bibitem[Ji et~al.(2023)Ji, Lee, Frieske, Yu, Su, Xu, Ishii, Bang, Madotto, and Fung]{ji2023survey}
Ji, Z., Lee, N., Frieske, R., Yu, T., Su, D., Xu, Y., Ishii, E., Bang, Y.~J., Madotto, A., and Fung, P.
\newblock Survey of hallucination in natural language generation.
\newblock \emph{ACM Computing Surveys}, 55\penalty0 (12):\penalty0 1--38, 2023.

\bibitem[Jiang et~al.(2023{\natexlab{a}})Jiang, Ren, and Lin]{jiang2023llm}
Jiang, D., Ren, X., and Lin, B.~Y.
\newblock Llm-blender: Ensembling large language models with pairwise ranking and generative fusion.
\newblock In \emph{Proceedings of the 61st Annual Meeting of the Association for Computational Linguistics (Volume 1: Long Papers)}, pp.\  14165--14178, 2023{\natexlab{a}}.

\bibitem[Jiang et~al.(2023{\natexlab{b}})Jiang, Xu, Gao, Sun, Liu, Dwivedi-Yu, Yang, Callan, and Neubig]{jiang2023active}
Jiang, Z., Xu, F.~F., Gao, L., Sun, Z., Liu, Q., Dwivedi-Yu, J., Yang, Y., Callan, J., and Neubig, G.
\newblock Active retrieval augmented generation.
\newblock In \emph{Proceedings of the 2023 Conference on Empirical Methods in Natural Language Processing}, pp.\  7969--7992, 2023{\natexlab{b}}.

\bibitem[Johnson(2006)]{johnson2006collaboration}
Johnson, J.~P.
\newblock Collaboration, peer review and open source software.
\newblock \emph{Information Economics and Policy}, 18\penalty0 (4):\penalty0 477--497, 2006.

\bibitem[Jung et~al.(2024)Jung, Brahman, and Choi]{jung2024trust}
Jung, J., Brahman, F., and Choi, Y.
\newblock Trust or escalate: Llm judges with provable guarantees for human agreement.
\newblock \emph{arXiv preprint arXiv:2407.18370}, 2024.

\bibitem[Kaplan et~al.(2020)Kaplan, McCandlish, Henighan, Brown, Chess, Child, Gray, Radford, Wu, and Amodei]{kaplan2020scaling}
Kaplan, J., McCandlish, S., Henighan, T., Brown, T.~B., Chess, B., Child, R., Gray, S., Radford, A., Wu, J., and Amodei, D.
\newblock Scaling laws for neural language models.
\newblock \emph{arXiv preprint arXiv:2001.08361}, 2020.

\bibitem[Karamolegkou et~al.(2023)Karamolegkou, Li, Zhou, and S{\o}gaard]{karamolegkou2023copyright}
Karamolegkou, A., Li, J., Zhou, L., and S{\o}gaard, A.
\newblock Copyright violations and large language models.
\newblock In \emph{The 2023 Conference on Empirical Methods in Natural Language Processing}, 2023.

\bibitem[Kasai et~al.(2024)Kasai, Sakaguchi, Le~Bras, Asai, Yu, Radev, Smith, Choi, Inui, et~al.]{kasai2024realtime}
Kasai, J., Sakaguchi, K., Le~Bras, R., Asai, A., Yu, X., Radev, D., Smith, N.~A., Choi, Y., Inui, K., et~al.
\newblock Realtime qa: what's the answer right now?
\newblock \emph{Advances in Neural Information Processing Systems}, 36, 2024.

\bibitem[Kirk et~al.(2024)Kirk, Whitefield, R{\"o}ttger, Bean, Margatina, Ciro, Mosquera, Bartolo, Williams, He, et~al.]{kirk2024prism}
Kirk, H.~R., Whitefield, A., R{\"o}ttger, P., Bean, A., Margatina, K., Ciro, J., Mosquera, R., Bartolo, M., Williams, A., He, H., et~al.
\newblock The prism alignment project: What participatory, representative and individualised human feedback reveals about the subjective and multicultural alignment of large language models.
\newblock \emph{arXiv preprint arXiv:2404.16019}, 2024.

\bibitem[Kirkpatrick et~al.(2017)Kirkpatrick, Pascanu, Rabinowitz, Veness, Desjardins, Rusu, Milan, Quan, Ramalho, Grabska-Barwinska, et~al.]{kirkpatrick2017overcoming}
Kirkpatrick, J., Pascanu, R., Rabinowitz, N., Veness, J., Desjardins, G., Rusu, A.~A., Milan, K., Quan, J., Ramalho, T., Grabska-Barwinska, A., et~al.
\newblock Overcoming catastrophic forgetting in neural networks.
\newblock \emph{Proceedings of the national academy of sciences}, 114\penalty0 (13):\penalty0 3521--3526, 2017.

\bibitem[Kumar et~al.(2023)Kumar, Balachandran, Njoo, Anastasopoulos, and Tsvetkov]{kumar2023language}
Kumar, S., Balachandran, V., Njoo, L., Anastasopoulos, A., and Tsvetkov, Y.
\newblock Language generation models can cause harm: So what can we do about it? an actionable survey.
\newblock In \emph{Proceedings of the 17th Conference of the European Chapter of the Association for Computational Linguistics}, pp.\  3299--3321, 2023.

\bibitem[Kumar et~al.(2024)Kumar, Park, Tsvetkov, Smith, and Hajishirzi]{kumar2024compo}
Kumar, S., Park, C.~Y., Tsvetkov, Y., Smith, N.~A., and Hajishirzi, H.
\newblock Compo: Community preferences for language model personalization.
\newblock \emph{arXiv preprint arXiv:2410.16027}, 2024.

\bibitem[Lazaridou et~al.(2021)Lazaridou, Kuncoro, Gribovskaya, Agrawal, Liska, Terzi, Gimenez, de~Masson~d’Autume, Ruder, Yogatama, et~al.]{lazaridou2021pitfalls}
Lazaridou, A., Kuncoro, A., Gribovskaya, E., Agrawal, D., Liska, A., Terzi, T., Gimenez, M., de~Masson~d’Autume, C., Ruder, S., Yogatama, D., et~al.
\newblock Pitfalls of static language modelling.
\newblock \emph{arXiv preprint arXiv:2102.01951}, 2021.

\bibitem[Leibo et~al.(2024)Leibo, Vezhnevets, Diaz, Agapiou, Cunningham, Sunehag, Haas, Koster, Du{\'e}{\~n}ez-Guzm{\'a}n, Isaac, et~al.]{leibo2024theory}
Leibo, J.~Z., Vezhnevets, A.~S., Diaz, M., Agapiou, J.~P., Cunningham, W.~A., Sunehag, P., Haas, J., Koster, R., Du{\'e}{\~n}ez-Guzm{\'a}n, E.~A., Isaac, W.~S., et~al.
\newblock A theory of appropriateness with applications to generative artificial intelligence.
\newblock \emph{arXiv preprint arXiv:2412.19010}, 2024.

\bibitem[Leng et~al.(2024)Leng, Zhang, Chen, Li, Lu, Miao, and Bing]{Leng_2024_CVPR}
Leng, S., Zhang, H., Chen, G., Li, X., Lu, S., Miao, C., and Bing, L.
\newblock Mitigating object hallucinations in large vision-language models through visual contrastive decoding.
\newblock In \emph{Proceedings of the IEEE/CVF Conference on Computer Vision and Pattern Recognition (CVPR)}, pp.\  13872--13882, June 2024.

\bibitem[Li et~al.(2022)Li, Gururangan, Dettmers, Lewis, Althoff, Smith, and Zettlemoyer]{li2022branch}
Li, M., Gururangan, S., Dettmers, T., Lewis, M., Althoff, T., Smith, N.~A., and Zettlemoyer, L.
\newblock Branch-train-merge: Embarrassingly parallel training of expert language models.
\newblock \emph{arXiv preprint arXiv:2208.03306}, 2022.

\bibitem[Li et~al.(2023)Li, Holtzman, Fried, Liang, Eisner, Hashimoto, Zettlemoyer, and Lewis]{li-etal-2023-contrastive}
Li, X.~L., Holtzman, A., Fried, D., Liang, P., Eisner, J., Hashimoto, T., Zettlemoyer, L., and Lewis, M.
\newblock Contrastive decoding: Open-ended text generation as optimization.
\newblock In \emph{Proceedings of the 61st Annual Meeting of the Association for Computational Linguistics (Volume 1: Long Papers)}, 2023.

\bibitem[Liang et~al.(2023{\natexlab{a}})Liang, Bommasani, Lee, Tsipras, Soylu, Yasunaga, Zhang, Narayanan, Wu, Kumar, et~al.]{liangholistic}
Liang, P., Bommasani, R., Lee, T., Tsipras, D., Soylu, D., Yasunaga, M., Zhang, Y., Narayanan, D., Wu, Y., Kumar, A., et~al.
\newblock Holistic evaluation of language models.
\newblock \emph{Transactions on Machine Learning Research}, 2023{\natexlab{a}}.

\bibitem[Liang et~al.(2023{\natexlab{b}})Liang, He, Jiao, Wang, Wang, Wang, Yang, Shi, and Tu]{liang2023encouraging}
Liang, T., He, Z., Jiao, W., Wang, X., Wang, Y., Wang, R., Yang, Y., Shi, S., and Tu, Z.
\newblock Encouraging divergent thinking in large language models through multi-agent debate.
\newblock \emph{arXiv preprint arXiv:2305.19118}, 2023{\natexlab{b}}.

\bibitem[Lin et~al.(2024)Lin, Lin, Xiong, Diao, Liu, Zhang, Pan, Wang, Hu, Zhang, et~al.]{lin2024mitigating}
Lin, Y., Lin, H., Xiong, W., Diao, S., Liu, J., Zhang, J., Pan, R., Wang, H., Hu, W., Zhang, H., et~al.
\newblock Mitigating the alignment tax of rlhf.
\newblock In \emph{Proceedings of the 2024 Conference on Empirical Methods in Natural Language Processing}, pp.\  580--606, 2024.

\bibitem[Liu et~al.(2021)Liu, Sap, Lu, Swayamdipta, Bhagavatula, Smith, and Choi]{liu-etal-2021-dexperts}
Liu, A., Sap, M., Lu, X., Swayamdipta, S., Bhagavatula, C., Smith, N.~A., and Choi, Y.
\newblock {DE}xperts: Decoding-time controlled text generation with experts and anti-experts.
\newblock In \emph{Proceedings of the 59th Annual Meeting of the Association for Computational Linguistics and the 11th International Joint Conference on Natural Language Processing (Volume 1: Long Papers)}, 2021.

\bibitem[Liu et~al.(2024)Liu, Han, Wang, Tsvetkov, Choi, and Smith]{liu-etal-2024-tuning}
Liu, A., Han, X., Wang, Y., Tsvetkov, Y., Choi, Y., and Smith, N.~A.
\newblock Tuning language models by proxy.
\newblock In \emph{First Conference on Language Modeling}, 2024.

\bibitem[Lu et~al.(2023)Lu, Yuan, Lin, Lin, Yuan, Zhou, and Zhou]{lu2023routing}
Lu, K., Yuan, H., Lin, R., Lin, J., Yuan, Z., Zhou, C., and Zhou, J.
\newblock Routing to the expert: Efficient reward-guided ensemble of large language models.
\newblock \emph{arXiv preprint arXiv:2311.08692}, 2023.

\bibitem[Luo et~al.(2023)Luo, Yang, Meng, Li, Zhou, and Zhang]{luo2023empirical}
Luo, Y., Yang, Z., Meng, F., Li, Y., Zhou, J., and Zhang, Y.
\newblock An empirical study of catastrophic forgetting in large language models during continual fine-tuning.
\newblock \emph{arXiv preprint arXiv:2308.08747}, 2023.

\bibitem[Madaan et~al.(2024)Madaan, Tandon, Gupta, Hallinan, Gao, Wiegreffe, Alon, Dziri, Prabhumoye, Yang, et~al.]{madaan2024self}
Madaan, A., Tandon, N., Gupta, P., Hallinan, S., Gao, L., Wiegreffe, S., Alon, U., Dziri, N., Prabhumoye, S., Yang, Y., et~al.
\newblock Self-refine: Iterative refinement with self-feedback.
\newblock \emph{Advances in Neural Information Processing Systems}, 36, 2024.

\bibitem[Mavromatis et~al.(2024)Mavromatis, Karypis, and Karypis]{mavromatis2024pack}
Mavromatis, C., Karypis, P., and Karypis, G.
\newblock Pack of {LLM}s: Model fusion at test-time via perplexity optimization.
\newblock In \emph{First Conference on Language Modeling}, 2024.

\bibitem[McCloskey \& Cohen(1989)McCloskey and Cohen]{mccloskey1989catastrophic}
McCloskey, M. and Cohen, N.~J.
\newblock Catastrophic interference in connectionist networks: The sequential learning problem.
\newblock In \emph{Psychology of learning and motivation}, volume~24, pp.\  109--165. Elsevier, 1989.

\bibitem[Meng et~al.(2022)Meng, Bau, Andonian, and Belinkov]{meng2022locating}
Meng, K., Bau, D., Andonian, A., and Belinkov, Y.
\newblock Locating and editing factual associations in {GPT}.
\newblock \emph{Advances in Neural Information Processing Systems}, 36, 2022.
\newblock arXiv:2202.05262.

\bibitem[Min et~al.(2024)Min, Gururangan, Wallace, Shi, Hajishirzi, Smith, and Zettlemoyer]{minsilo}
Min, S., Gururangan, S., Wallace, E., Shi, W., Hajishirzi, H., Smith, N.~A., and Zettlemoyer, L.
\newblock Silo language models: Isolating legal risk in a nonparametric datastore.
\newblock In \emph{The Twelfth International Conference on Learning Representations}, 2024.

\bibitem[Mireshghallah et~al.(2024)Mireshghallah, Kim, Zhou, Tsvetkov, Sap, Shokri, and Choi]{mireshghallahcan}
Mireshghallah, N., Kim, H., Zhou, X., Tsvetkov, Y., Sap, M., Shokri, R., and Choi, Y.
\newblock Can llms keep a secret? testing privacy implications of language models via contextual integrity theory.
\newblock In \emph{The Twelfth International Conference on Learning Representations}, 2024.

\bibitem[Mitchell et~al.(2024)Mitchell, Rafailov, Sharma, Finn, and Manning]{mitchell2024an}
Mitchell, E., Rafailov, R., Sharma, A., Finn, C., and Manning, C.~D.
\newblock An emulator for fine-tuning large language models using small language models.
\newblock In \emph{The Twelfth International Conference on Learning Representations}, 2024.

\bibitem[Muqeeth et~al.(2024)Muqeeth, Liu, Liu, and Raffel]{muqeethlearning}
Muqeeth, M., Liu, H., Liu, Y., and Raffel, C.
\newblock Learning to route among specialized experts for zero-shot generalization.
\newblock In \emph{Forty-first International Conference on Machine Learning}, 2024.

\bibitem[Nanda et~al.(2023)Nanda, Chan, Lieberum, Smith, and Steinhardt]{nandaprogress}
Nanda, N., Chan, L., Lieberum, T., Smith, J., and Steinhardt, J.
\newblock Progress measures for grokking via mechanistic interpretability.
\newblock In \emph{The Eleventh International Conference on Learning Representations}, 2023.

\bibitem[Naous et~al.(2023)Naous, Ryan, Ritter, and Xu]{naous2023having}
Naous, T., Ryan, M.~J., Ritter, A., and Xu, W.
\newblock Having beer after prayer? measuring cultural bias in large language models.
\newblock \emph{arXiv preprint arXiv:2305.14456}, 2023.

\bibitem[Ong et~al.(2024)Ong, Almahairi, Wu, Chiang, Wu, Gonzalez, Kadous, and Stoica]{ong2024routellm}
Ong, I., Almahairi, A., Wu, V., Chiang, W.-L., Wu, T., Gonzalez, J.~E., Kadous, M.~W., and Stoica, I.
\newblock Routellm: Learning to route llms with preference data.
\newblock \emph{arXiv preprint arXiv:2406.18665}, 2024.

\bibitem[Pei et~al.(2023)Pei, Yang, and Klein]{pei-etal-2023-preadd}
Pei, J., Yang, K., and Klein, D.
\newblock {PREADD}: Prefix-adaptive decoding for controlled text generation.
\newblock In \emph{Findings of the Association for Computational Linguistics: ACL 2023}, 2023.

\bibitem[Pfeiffer et~al.(2020)Pfeiffer, Kamath, R{\"u}ckl{\'e}, Cho, and Gurevych]{pfeiffer2020adapterfusion}
Pfeiffer, J., Kamath, A., R{\"u}ckl{\'e}, A., Cho, K., and Gurevych, I.
\newblock Adapterfusion: Non-destructive task composition for transfer learning.
\newblock \emph{arXiv preprint arXiv:2005.00247}, 2020.

\bibitem[Press et~al.(2022)Press, Zhang, Min, Schmidt, Smith, and Lewis]{Press2022MeasuringAN}
Press, O., Zhang, M., Min, S., Schmidt, L., Smith, N.~A., and Lewis, M.
\newblock Measuring and narrowing the compositionality gap in language models.
\newblock \emph{ArXiv}, abs/2210.03350, 2022.

\bibitem[Raffel(2021)]{raffel2021call}
Raffel, C.
\newblock A call to build models like we build open-source software, 2021.
\newblock Accessed: 2025-01-15.

\bibitem[Rao et~al.(2024)Rao, Yerukola, Shah, Reinecke, and Sap]{rao2024normad}
Rao, A., Yerukola, A., Shah, V., Reinecke, K., and Sap, M.
\newblock Normad: A benchmark for measuring the cultural adaptability of large language models.
\newblock \emph{arXiv preprint arXiv:2404.12464}, 2024.

\bibitem[Santurkar et~al.(2023)Santurkar, Durmus, Ladhak, Lee, Liang, and Hashimoto]{santurkar2023whose}
Santurkar, S., Durmus, E., Ladhak, F., Lee, C., Liang, P., and Hashimoto, T.
\newblock Whose opinions do language models reflect?
\newblock In \emph{International Conference on Machine Learning}, pp.\  29971--30004. PMLR, 2023.

\bibitem[Sennrich et~al.(2024)Sennrich, Vamvas, and Mohammadshahi]{sennrich-etal-2024-mitigating}
Sennrich, R., Vamvas, J., and Mohammadshahi, A.
\newblock Mitigating hallucinations and off-target machine translation with source-contrastive and language-contrastive decoding.
\newblock In \emph{Proceedings of the 18th Conference of the European Chapter of the Association for Computational Linguistics (Volume 2: Short Papers)}, 2024.

\bibitem[Shen et~al.(2024)Shen, Lang, Wang, Kim, and Sontag]{shen-etal-2024-learning}
Shen, Z., Lang, H., Wang, B., Kim, Y., and Sontag, D.
\newblock Learning to decode collaboratively with multiple language models.
\newblock In \emph{Proceedings of the 62nd Annual Meeting of the Association for Computational Linguistics (Volume 1: Long Papers)}, 2024.

\bibitem[Shi et~al.(2024{\natexlab{a}})Shi, Chen, Hu, Liu, Hajishirzi, Smith, and Du]{shi-etal-2024-decoding}
Shi, R., Chen, Y., Hu, Y., Liu, A., Hajishirzi, H., Smith, N.~A., and Du, S.~S.
\newblock Decoding-time language model alignment with multiple objectives.
\newblock In \emph{The Thirty-eighth Annual Conference on Neural Information Processing Systems}, 2024{\natexlab{a}}.

\bibitem[Shi et~al.(2023)Shi, Min, Yasunaga, Seo, James, Lewis, Zettlemoyer, and Yih]{shi2023replug}
Shi, W., Min, S., Yasunaga, M., Seo, M., James, R., Lewis, M., Zettlemoyer, L., and Yih, W.-t.
\newblock Replug: Retrieval-augmented black-box language models.
\newblock \emph{arXiv preprint arXiv:2301.12652}, 2023.

\bibitem[Shi et~al.(2024{\natexlab{b}})Shi, Han, Lewis, Tsvetkov, Zettlemoyer, and Yih]{shi-etal-2024-trusting}
Shi, W., Han, X., Lewis, M., Tsvetkov, Y., Zettlemoyer, L., and Yih, W.-t.
\newblock Trusting your evidence: Hallucinate less with context-aware decoding.
\newblock In \emph{Proceedings of the 2024 Conference of the North American Chapter of the Association for Computational Linguistics: Human Language Technologies (Volume 2: Short Papers)}, 2024{\natexlab{b}}.

\bibitem[Shi et~al.(2024{\natexlab{c}})Shi, Li, Zhang, Ziems, Horesh, de~Paula, Yang, et~al.]{shi2024culturebank}
Shi, W., Li, R., Zhang, Y., Ziems, C., Horesh, R., de~Paula, R.~A., Yang, D., et~al.
\newblock Culturebank: An online community-driven knowledge base towards culturally aware language technologies.
\newblock \emph{arXiv preprint arXiv:2404.15238}, 2024{\natexlab{c}}.

\bibitem[Shinn et~al.(2024)Shinn, Cassano, Gopinath, Narasimhan, and Yao]{shinn2024reflexion}
Shinn, N., Cassano, F., Gopinath, A., Narasimhan, K., and Yao, S.
\newblock Reflexion: Language agents with verbal reinforcement learning.
\newblock \emph{Advances in Neural Information Processing Systems}, 36, 2024.

\bibitem[Shnitzer et~al.(2023)Shnitzer, Ou, Silva, Soule, Sun, Solomon, Thompson, and Yurochkin]{shnitzer2023large}
Shnitzer, T., Ou, A., Silva, M., Soule, K., Sun, Y., Solomon, J., Thompson, N., and Yurochkin, M.
\newblock Large language model routing with benchmark datasets.
\newblock \emph{arXiv preprint arXiv:2309.15789}, 2023.

\bibitem[Song et~al.(2023)Song, Khanuja, Liu, Faisal, Ostapenko, Winata, Aji, Cahyawijaya, Tsvetkov, Anastasopoulos, et~al.]{song2023globalbench}
Song, Y., Khanuja, S., Liu, P., Faisal, F., Ostapenko, A., Winata, G., Aji, A., Cahyawijaya, S., Tsvetkov, Y., Anastasopoulos, A., et~al.
\newblock Globalbench: A benchmark for global progress in natural language processing.
\newblock In \emph{Proceedings of the 2023 Conference on Empirical Methods in Natural Language Processing}, pp.\  14157--14171, 2023.

\bibitem[Sorensen et~al.(2024{\natexlab{a}})Sorensen, Jiang, Hwang, Levine, Pyatkin, West, Dziri, Lu, Rao, Bhagavatula, et~al.]{sorensen2024value}
Sorensen, T., Jiang, L., Hwang, J.~D., Levine, S., Pyatkin, V., West, P., Dziri, N., Lu, X., Rao, K., Bhagavatula, C., et~al.
\newblock Value kaleidoscope: Engaging ai with pluralistic human values, rights, and duties.
\newblock In \emph{Proceedings of the AAAI Conference on Artificial Intelligence}, volume~38, pp.\  19937--19947, 2024{\natexlab{a}}.

\bibitem[Sorensen et~al.(2024{\natexlab{b}})Sorensen, Moore, Fisher, Gordon, Mireshghallah, Rytting, Ye, Jiang, Lu, Dziri, et~al.]{sorensenposition}
Sorensen, T., Moore, J., Fisher, J., Gordon, M.~L., Mireshghallah, N., Rytting, C.~M., Ye, A., Jiang, L., Lu, X., Dziri, N., et~al.
\newblock Position: A roadmap to pluralistic alignment.
\newblock In \emph{Forty-first International Conference on Machine Learning}, 2024{\natexlab{b}}.

\bibitem[Sprague et~al.(2024)Sprague, Yin, Rodriguez, Jiang, Wadhwa, Singhal, Zhao, Ye, Mahowald, and Durrett]{sprague2024cot}
Sprague, Z., Yin, F., Rodriguez, J.~D., Jiang, D., Wadhwa, M., Singhal, P., Zhao, X., Ye, X., Mahowald, K., and Durrett, G.
\newblock To cot or not to cot? chain-of-thought helps mainly on math and symbolic reasoning.
\newblock \emph{arXiv preprint arXiv:2409.12183}, 2024.

\bibitem[Stolfo et~al.(2023)Stolfo, Belinkov, and Sachan]{stolfo-etal-2023-mechanistic}
Stolfo, A., Belinkov, Y., and Sachan, M.
\newblock A mechanistic interpretation of arithmetic reasoning in language models using causal mediation analysis.
\newblock In \emph{Proceedings of the 2023 Conference on Empirical Methods in Natural Language Processing}, 2023.

\bibitem[Stripelis et~al.(2024)Stripelis, Hu, Zhang, Xu, Shah, Jin, Yao, Avestimehr, and He]{stripelis2024tensoropera}
Stripelis, D., Hu, Z., Zhang, J., Xu, Z., Shah, A.~D., Jin, H., Yao, Y., Avestimehr, S., and He, C.
\newblock Tensoropera router: A multi-model router for efficient llm inference.
\newblock \emph{arXiv preprint arXiv:2408.12320}, 2024.

\bibitem[Subramaniam et~al.(2024)Subramaniam, Torralba, and Li]{subramaniam2024debategpt}
Subramaniam, V., Torralba, A., and Li, S.
\newblock Debategpt: Fine-tuning large language models with multi-agent debate supervision.
\newblock 2024.

\bibitem[Subramaniam et~al.(2025)Subramaniam, Du, Tenenbaum, Torralba, Li, and Mordatch]{subramaniam2025multiagent}
Subramaniam, V., Du, Y., Tenenbaum, J.~B., Torralba, A., Li, S., and Mordatch, I.
\newblock Multiagent finetuning: Self improvement with diverse reasoning chains.
\newblock \emph{arXiv preprint arXiv:2501.05707}, 2025.

\bibitem[Sukhbaatar et~al.(2024)Sukhbaatar, Golovneva, Sharma, Xu, Lin, Rozi{\`e}re, Kahn, Li, Yih, Weston, et~al.]{sukhbaatar2024branch}
Sukhbaatar, S., Golovneva, O., Sharma, V., Xu, H., Lin, X.~V., Rozi{\`e}re, B., Kahn, J., Li, D., Yih, W.-t., Weston, J., et~al.
\newblock Branch-train-mix: Mixing expert llms into a mixture-of-experts llm.
\newblock \emph{arXiv preprint arXiv:2403.07816}, 2024.

\bibitem[Sun et~al.(2022)Sun, Liu, Qiu, and Huang]{sun2022paradigm}
Sun, T.-X., Liu, X.-Y., Qiu, X.-P., and Huang, X.-J.
\newblock Paradigm shift in natural language processing.
\newblock \emph{Machine Intelligence Research}, 19\penalty0 (3):\penalty0 169--183, 2022.

\bibitem[Suzgun et~al.(2023)Suzgun, Scales, Sch{\"a}rli, Gehrmann, Tay, Chung, Chowdhery, Le, Chi, Zhou, et~al.]{suzgun2023challenging}
Suzgun, M., Scales, N., Sch{\"a}rli, N., Gehrmann, S., Tay, Y., Chung, H.~W., Chowdhery, A., Le, Q., Chi, E., Zhou, D., et~al.
\newblock Challenging big-bench tasks and whether chain-of-thought can solve them.
\newblock In \emph{Findings of the Association for Computational Linguistics: ACL 2023}, pp.\  13003--13051, 2023.

\bibitem[Tambon et~al.(2024)Tambon, Nikanjam, Khomh, and Antoniol]{tambon2024assessing}
Tambon, F., Nikanjam, A., Khomh, F., and Antoniol, G.
\newblock Assessing programming task difficulty for efficient evaluation of large language models.
\newblock \emph{arXiv preprint arXiv:2407.21227}, 2024.

\bibitem[Team et~al.(2023)Team, Anil, Borgeaud, Alayrac, Yu, Soricut, Schalkwyk, Dai, Hauth, Millican, et~al.]{team2023gemini}
Team, G., Anil, R., Borgeaud, S., Alayrac, J.-B., Yu, J., Soricut, R., Schalkwyk, J., Dai, A.~M., Hauth, A., Millican, K., et~al.
\newblock Gemini: a family of highly capable multimodal models.
\newblock \emph{arXiv preprint arXiv:2312.11805}, 2023.

\bibitem[Team(2023)]{team2023internlm}
Team, I.
\newblock Internlm: A multilingual language model with progressively enhanced capabilities, 2023.

\bibitem[Vaswani(2017)]{vaswani2017attention}
Vaswani, A.
\newblock Attention is all you need.
\newblock \emph{Advances in Neural Information Processing Systems}, 2017.

\bibitem[Wang et~al.(2024)Wang, Polo, Sun, Kundu, Xing, and Yurochkin]{wangfusing}
Wang, H., Polo, F.~M., Sun, Y., Kundu, S., Xing, E., and Yurochkin, M.
\newblock Fusing models with complementary expertise.
\newblock In \emph{The Twelfth International Conference on Learning Representations}, 2024.

\bibitem[Wang et~al.(2022{\natexlab{a}})Wang, Wei, Schuurmans, Le, Chi, Narang, Chowdhery, and Zhou]{wang2022self}
Wang, X., Wei, J., Schuurmans, D., Le, Q., Chi, E., Narang, S., Chowdhery, A., and Zhou, D.
\newblock Self-consistency improves chain of thought reasoning in language models.
\newblock \emph{arXiv preprint arXiv:2203.11171}, 2022{\natexlab{a}}.

\bibitem[Wang et~al.(2022{\natexlab{b}})Wang, Agarwal, Mukherjee, Liu, Gao, Awadallah, and Gao]{wang2022adamixmixtureofadaptationsparameterefficientmodel}
Wang, Y., Agarwal, S., Mukherjee, S., Liu, X., Gao, J., Awadallah, A.~H., and Gao, J.
\newblock Adamix: Mixture-of-adaptations for parameter-efficient model tuning, 2022{\natexlab{b}}.

\bibitem[Wei et~al.(2024)Wei, Shi, Huang, Smith, Zhang, Zettlemoyer, Li, and Henderson]{weievaluating}
Wei, B., Shi, W., Huang, Y., Smith, N.~A., Zhang, C., Zettlemoyer, L., Li, K., and Henderson, P.
\newblock Evaluating copyright takedown methods for language models.
\newblock In \emph{The Thirty-eight Conference on Neural Information Processing Systems Datasets and Benchmarks Track}, 2024.

\bibitem[Wu et~al.(2024{\natexlab{a}})Wu, Bansal, Zhang, Wu, Li, Zhu, Jiang, Zhang, Zhang, Liu, et~al.]{wu2024autogen}
Wu, Q., Bansal, G., Zhang, J., Wu, Y., Li, B., Zhu, E., Jiang, L., Zhang, X., Zhang, S., Liu, J., et~al.
\newblock Autogen: Enabling next-gen llm applications via multi-agent conversation.
\newblock In \emph{ICLR 2024 Workshop on Large Language Model (LLM) Agents}, 2024{\natexlab{a}}.

\bibitem[Wu et~al.(2024{\natexlab{b}})Wu, Huang, and Wei]{wu2024mixture}
Wu, X., Huang, S., and Wei, F.
\newblock Mixture of lora experts.
\newblock \emph{arXiv preprint arXiv:2404.13628}, 2024{\natexlab{b}}.

\bibitem[Xiong et~al.(2023)Xiong, Ding, Cao, Liu, and Qin]{xiong2023examining}
Xiong, K., Ding, X., Cao, Y., Liu, T., and Qin, B.
\newblock Examining inter-consistency of large language models collaboration: An in-depth analysis via debate.
\newblock \emph{arXiv preprint arXiv:2305.11595}, 2023.

\bibitem[Xu et~al.(2023)Xu, Sun, Zheng, Geng, Zhao, Feng, Tao, and Jiang]{xu2023wizardlm}
Xu, C., Sun, Q., Zheng, K., Geng, X., Zhao, P., Feng, J., Tao, C., and Jiang, D.
\newblock Wizardlm: Empowering large language models to follow complex instructions.
\newblock \emph{arXiv preprint arXiv:2304.12244}, 2023.

\bibitem[Xu et~al.(2024)Xu, Wu, Diao, Liu, Wang, Chen, and Gao]{xu2024sayself}
Xu, T., Wu, S., Diao, S., Liu, X., Wang, X., Chen, Y., and Gao, J.
\newblock Sayself: Teaching llms to express confidence with self-reflective rationales.
\newblock \emph{arXiv preprint arXiv:2405.20974}, 2024.

\bibitem[Yadav et~al.(2024{\natexlab{a}})Yadav, Raffel, Muqeeth, Caccia, Liu, Chen, Bansal, Choshen, and Sordoni]{yadav2024survey}
Yadav, P., Raffel, C., Muqeeth, M., Caccia, L., Liu, H., Chen, T., Bansal, M., Choshen, L., and Sordoni, A.
\newblock A survey on model moerging: Recycling and routing among specialized experts for collaborative learning.
\newblock \emph{arXiv preprint arXiv:2408.07057}, 2024{\natexlab{a}}.

\bibitem[Yadav et~al.(2024{\natexlab{b}})Yadav, Tam, Choshen, Raffel, and Bansal]{yadav2024ties}
Yadav, P., Tam, D., Choshen, L., Raffel, C.~A., and Bansal, M.
\newblock Ties-merging: Resolving interference when merging models.
\newblock \emph{Advances in Neural Information Processing Systems}, 36, 2024{\natexlab{b}}.

\bibitem[Yang et~al.(2024)Yang, Gribovskaya, Kassner, Geva, and Riedel]{yang-etal-2024-large-language-models}
Yang, S., Gribovskaya, E., Kassner, N., Geva, M., and Riedel, S.
\newblock Do large language models latently perform multi-hop reasoning?
\newblock In \emph{Proceedings of the 62nd Annual Meeting of the Association for Computational Linguistics (Volume 1: Long Papers)}, 2024.

\bibitem[Yao et~al.(2024)Yao, Duan, Xu, Cai, Sun, and Zhang]{yao2024survey}
Yao, Y., Duan, J., Xu, K., Cai, Y., Sun, Z., and Zhang, Y.
\newblock A survey on large language model (llm) security and privacy: The good, the bad, and the ugly.
\newblock \emph{High-Confidence Computing}, pp.\  100211, 2024.

\bibitem[Yin et~al.(2024)Yin, Ye, and Durrett]{yin2024lofit}
Yin, F., Ye, X., and Durrett, G.
\newblock Lofit: Localized fine-tuning on llm representations.
\newblock \emph{arXiv preprint arXiv:2406.01563}, 2024.

\bibitem[Yu et~al.(2024)Yu, Yu, Yu, Huang, and Li]{yu2024language}
Yu, L., Yu, B., Yu, H., Huang, F., and Li, Y.
\newblock Language models are super mario: Absorbing abilities from homologous models as a free lunch.
\newblock In \emph{Forty-first International Conference on Machine Learning}, 2024.

\bibitem[Yue et~al.(2023)Yue, Zhao, Zhang, Du, and Yao]{yue2023large}
Yue, M., Zhao, J., Zhang, M., Du, L., and Yao, Z.
\newblock Large language model cascades with mixture of thoughts representations for cost-efficient reasoning.
\newblock \emph{arXiv preprint arXiv:2310.03094}, 2023.

\bibitem[Zhang et~al.(2024)Zhang, Wang, Hua, Qi, Ding, and Zhou]{zhang2024cogenesis}
Zhang, K., Wang, J., Hua, E., Qi, B., Ding, N., and Zhou, B.
\newblock Cogenesis: A framework collaborating large and small language models for secure context-aware instruction following.
\newblock \emph{arXiv preprint arXiv:2403.03129}, 2024.

\bibitem[Zhang et~al.(2023)Zhang, Dong, Li, Zhang, Sun, Wang, Li, Hu, Zhang, Wu, et~al.]{zhang2023instruction}
Zhang, S., Dong, L., Li, X., Zhang, S., Sun, X., Wang, S., Li, J., Hu, R., Zhang, T., Wu, F., et~al.
\newblock Instruction tuning for large language models: A survey.
\newblock \emph{arXiv preprint arXiv:2308.10792}, 2023.

\bibitem[Zhao et~al.(2024{\natexlab{a}})Zhao, Plaza-del Arco, and Curry]{zhao2024language}
Zhao, J., Plaza-del Arco, F.~M., and Curry, A.~C.
\newblock Language model council: Benchmarking foundation models on highly subjective tasks by consensus.
\newblock \emph{arXiv preprint arXiv:2406.08598}, 2024{\natexlab{a}}.

\bibitem[Zhao et~al.(2024{\natexlab{b}})Zhao, Wang, Zhang, Jin, Zhu, Chen, and Xie]{zhaocompeteai}
Zhao, Q., Wang, J., Zhang, Y., Jin, Y., Zhu, K., Chen, H., and Xie, X.
\newblock Competeai: Understanding the competition dynamics of large language model-based agents.
\newblock In \emph{Forty-first International Conference on Machine Learning}, 2024{\natexlab{b}}.

\end{thebibliography}
\bibliographystyle{icml2025}

\end{document}